\documentclass{article}

\usepackage{microtype}
\usepackage{graphicx}
\usepackage{subfigure}
\usepackage{booktabs} %
\usepackage[dvipsnames]{xcolor}
\usepackage{xcolor,colortbl}
\usepackage{hyperref}

\usepackage[accepted]{icml2024}

\usepackage{amsmath}
\usepackage{amssymb}
\usepackage{mathtools}
\usepackage{amsthm}
\usepackage{multicol}
\usepackage{multirow}
\usepackage[capitalize,noabbrev]{cleveref}
\def\etal{\textit{et al}.}
\def\ie{\textit{i.e.}}
\def\eg{\textit{e.g.}}
\definecolor{mygray}{gray}{.9}
\theoremstyle{plain}

\theoremstyle{definition}

\theoremstyle{remark}

\newcommand{\app}{\raise.17ex\hbox{$\scriptstyle\sim$}}

\usepackage[textsize=tiny]{todonotes}

\usepackage{xspace}
\makeatletter\renewcommand\paragraph{\@startsection{paragraph}{4}{\z@}
	{0.05em \@plus1ex \@minus.2ex}{-.5em}{\normalfont\normalsize\bfseries}}\makeatother
\icmltitlerunning{D-iGPT}

\begin{document}

\twocolumn[
\icmltitle{Rejuvenating image-GPT as Strong Visual Representation Learners}

\icmlsetsymbol{equal}{*}

\begin{icmlauthorlist}
\icmlauthor{Sucheng Ren}{equal,1}
\icmlauthor{Zeyu Wang}{equal,2}
\icmlauthor{Hongru Zhu}{1}
\icmlauthor{Junfei Xiao}{1}
\icmlauthor{Alan Yuille}{1}
\icmlauthor{Cihang Xie}{2}
\end{icmlauthorlist}

\icmlaffiliation{1}{Johns Hopkins University}
\icmlaffiliation{2}{UC Santa Cruz}

\icmlcorrespondingauthor{Cihang Xie}{cixie@ucsc.edu}
\icmlkeywords{Machine Learning, ICML}

\vskip 0.3in
]

\printAffiliationsAndNotice{\icmlEqualContribution} %

\begin{abstract}
This paper enhances image-GPT (iGPT), one of the pioneering works that introduce autoregressive pretraining to predict the next pixels for visual representation learning.
    Two simple yet essential changes are made. First, we shift the prediction target from raw pixels to semantic tokens,  enabling a higher-level understanding of visual content. Second, we supplement the autoregressive modeling by instructing the model to predict not only the next tokens but also the visible tokens.
    This pipeline is particularly effective when semantic tokens are encoded by discriminatively trained models, such as CLIP. 
    We introduce this novel approach as D-iGPT. Extensive experiments showcase that D-iGPT excels as a strong learner of visual representations: A notable achievement is its compelling performance on the ImageNet-1K dataset --- by training on publicly available datasets, D-iGPT unprecedentedly achieves \textbf{90.0\%} top-1 accuracy with a vanilla ViT-H. Additionally, D-iGPT shows strong generalization on the downstream task.
    Code is available at \href{https://github.com/OliverRensu/D-iGPT}{https://github.com/OliverRensu/D-iGPT}.
\end{abstract}

\section{Introduction}
\begin{figure}
    \centering
         \includegraphics[width=\linewidth]{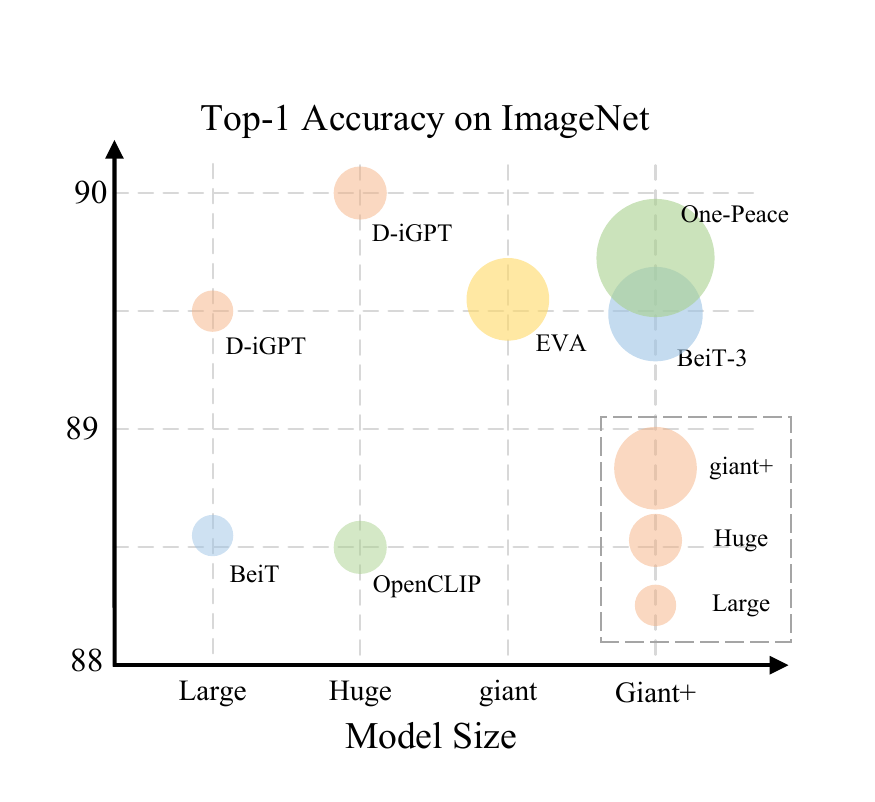}  \\
    \vspace{-.5em}
    \caption{ImageNet performance of models trained on publicly available datasets. We note that D-iGPT with ViT-H achieves the best performance, \ie, 90.0\% top-1 accuracy.}
    \vspace{-1.2em}
    \label{fig:acc}
\end{figure}

The advent of Large Language Models (LLMs)~\cite{gpt4,thoppilan2022lamda,touvron2023llama}, such as GPT series~\cite{gpt,gpt3,gpt4}, has catalyzed a transformative era in natural language processing (NLP), establishing new precedents for performance across a range of linguistic tasks. One of the key driving forces behind this tremendous success is autoregressive pretraining, which trains models to predict the most probable next tokens in a sequence. This strategy enables models to internalize a complex interplay of syntax and semantics, which in turn translates to their extraordinary prowess to process language with human-like capabilities.

Beyond NLP, autoregressive pretraining has also been a significant contributor in the field of computer vision. The pioneering model in this context is PixelCNN \cite{van2016pixel}, a deep autoregressive model designed to model the discrete probability of the raw pixel values and encode the complete set of dependencies in the image.  Building upon this foundation, image GPT (iGPT) \cite{igpt} represents a significant advancement, utilizing the flexible Transformer architecture \cite{vaswani2017attention} at a notably larger computational scale. iGPT's achievements are remarkable: it not only learned state-of-the-art visual representation for lower-resolution datasets such as CIFAR-10 but also demonstrated competitive performance on more complex datasets like ImageNet \cite{imagenet}.

Intriguingly, despite the initial successes of autoregressive pretraining in computer vision, recent trends have witnessed a rapid paradigm shift towards BERT-style pretraining \cite{bert}. This transition is significant, particularly when considering iGPT's initial findings of comparable performance between autoregressive and BERT-style pretraining in various tasks. Subsequent research, however, has increasingly favored BERT-style pretraining \cite{BEiT,mae} for its superior efficacy in visual representation learning. For example, MAE \cite{mae} demonstrates that simply predicting the values of randomly masked pixels can effectively serve as a scalable solution for visual representation learning.

In this paper, we revisit iGPT, challenging that \textit{autoregressive pretraining is actually capable of building strong vision learners, especially at scale}. Our methodology incorporates two critical modifications. First, acknowledging that images are inherently noisy and redundant, we follow BEiT \cite{BEiT} to ``tokenize'' images into semantic tokens. This adjustment reorients the autoregressive prediction focus from pixels to semantic tokens, thereby enabling a more nuanced understanding of the interplay among different image regions. Second, we complement the generative decoder, which is responsible for autoregressively predicting the next semantic token, with a discriminative decoder. This additional component is tasked with predicting the semantic tokens of the visible pixels. Moreover, an intriguing observation is that this pretraining pipeline works best when the semantic visual tokens are derived from models trained discriminatively, such as CLIP \cite{clip}. 
We term this enhanced approach as D-iGPT.

Extensive experiments across various datasets and tasks confirm the effectiveness of our proposed D-iGPT. With ImageNet-1K as the sole pertaining dataset, our base-size model achieves an 86.2\% top-1 classification accuracy, surpassing previous state-of-the-art by 0.6\%. By further scaling to the larger ImageNet-21K dataset, our huge-size model unprecedentedly achieves a \textbf{90.0\%} top-1 classification accuracy, outperforming all existing solutions developed using public datasets. We hope this work can catalyze the community to reevaluate the potential of autoregressive pretraining for visual representation learning.

\section{Related Work}
\subsection{Self-supervised Learning}
According to learning targets, self-supervised learning can be labeled as discriminative-based or generative-based. 

\paragraph{Discriminative Self-supervised Learning.} This paradigm focuses on learning transferable representation by defining a pre-task that scores the discriminative power of learned representations.  A notable strategy within this category is contrastive learning, which utilizes a contrastive loss to learn representation similarity or dissimilarity between the same images with different augmentation or entirely different images. For instance, Wu \etal~\cite{wu2018unsupervised} introduces instance discrimination, constructing positive and negative query-key pairs from the same or different images. SimCLR~\cite{simclr} further improves the performance with a projection head, strong data augmentations, and large-batch-size training. MoCo~\cite{moco,mocov2} incorporates a memory bank and a momentum encoder without the need for large batch sizes. CLIP~\cite{clip} extends this concept by incorporating language supervision through image-text pairings.

\paragraph{Generatieve Self-supervised Learning.}
In contrast to the discriminative approaches, generative self-supervised learning emphasizes training models to reconstruct the original inputs from corrupted versions.

Masked image modeling, inspired by BERT~\cite{bert} in NLP, is the dominant strategy in this line of research. For example, the pioneering work BEiT~\cite{BEiT} pretrains models to recover the corresponding semantic tokens based on the corrupted image patches. Other significant methods include MAE~\cite{mae}, SimMIM~\cite{simmim}, MaskFeat~\cite{maskfeat}, PeCo~\cite{peco}, MILAN~\cite{MILAN}, DeepMIM~\cite{deepmim}.

This study pivots towards a distinct facet of generative self-supervised learning, namely, autoregressive pretraining. In NLP, autoregressive pretraining is also highly regarded alongside BERT-style methods, especially effective in the era of LLMs~\cite{gpt4,touvron2023llama}. However, its progress in computer vision has not yet paralleled the heightened interest sparked by the initial success of iGPT \cite{igpt}. This paper aims to bridge this gap. We demonstrate that, with simple yet essential modification, autoregressive pretraining exhibits extraordinary capabilities in building strong vision models.

\subsection{ImageNet-1K Winning Solutions}

The advancements in ImageNet-1K performance have seen a significant boost, primarily driven by scaling datasets and model sizes. Liu \etal~\cite{liu2022swin} exemplify this trend with the successful training of SwinV2-G, a model equipped with \app3 billion parameters, using techniques like residual-post-norm and scaled cosine attention. Similarly, Dehghani \etal~\cite{dehghani2023scaling} have shown the impressive capabilities of ViT-22B, highlighting the feasibility of ``LLM-like'' scaling in computer vision. Zhang \etal~\cite{zhai2022scaling} investigate scaling both model and data, providing valuable insights into the interplay between scaling factors and performance. Another noteworthy development is by Chen \etal~\cite{chen2023symbolic} which discovers deep neural network training algorithms through program search, leading to the creation of the effective and memory-efficient optimizer Lion. However, a common limitation across these methods is their heavy reliance on private, in-house data, such as JFT-3B~\cite{zhai2022scaling}, which raises significant reproducibility concerns.

In contrast to the approaches above, there is a notable trend of employing public datasets to train more powerful vision models. For instance, Wang \etal~\cite{BEiT3} scale BEiT-3 to \app1.9 billion parameters using a combination of images, texts, and image-text pairs, all sourced from public datasets. Likewise, Fang \etal \cite{eva} successfully scaled up EVA, a vanilla ViT with \app1 billion parameters, using a total of \app29.6 million public images. One-Peace~\cite{wang2023one} presents a 4-billion-parameter model capable of unifying vision, audio, and language representations.
Our D-iGPT model stands out in this landscape by achieving superior performance than EVA and One-Peace, and meanwhile using smaller model and data sizes.

\begin{figure*}[t!]
    \centering
    \includegraphics[width=\linewidth]{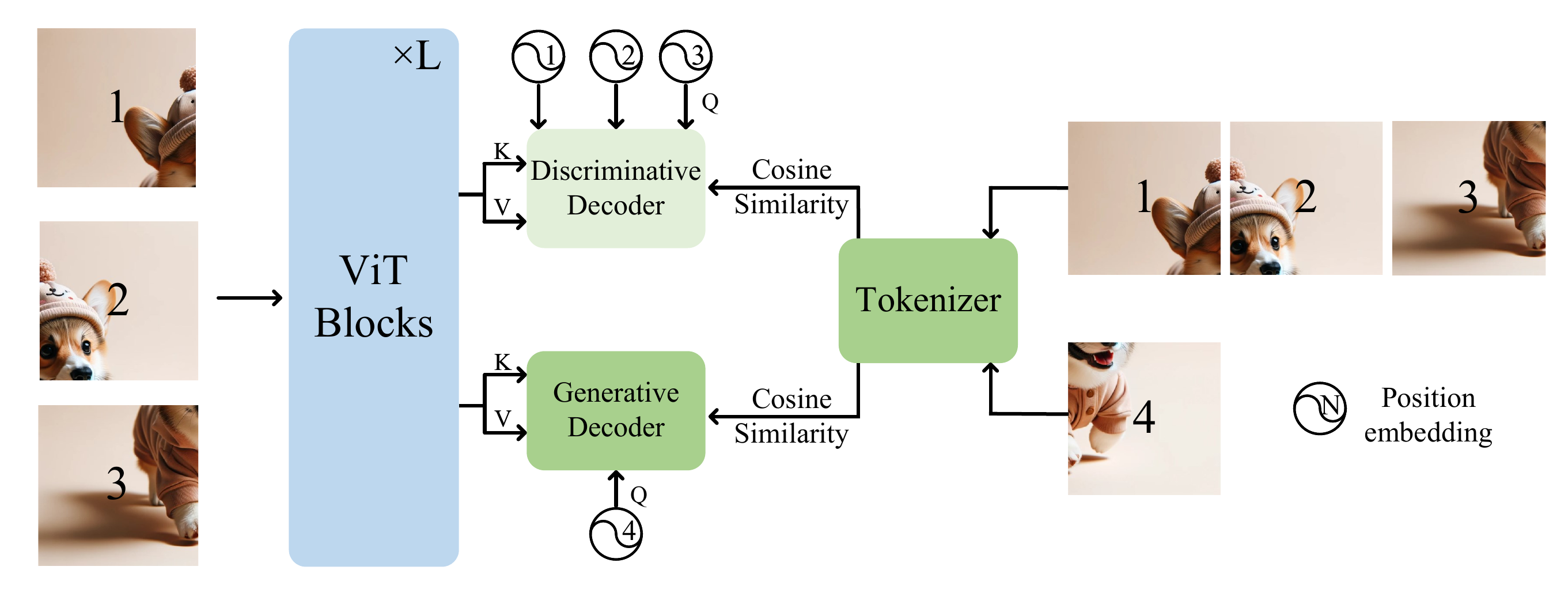}
    \vspace{-2em}
    \caption{The overview illustration of D-iGPT.}
    \vspace{-1em}
    \label{fig:method}
\end{figure*}

\section{Method}

We hereby first revisit iGPT in Section \ref{subsec:igpt}. Next, we present our enhanced version, D-iGPT, in Section \ref{subsec:digpt}, which shifts the prediction target from raw pixels to semantic tokens and additionally supplies supervision on visible tokens. Lastly,  the specifics of our model's architecture, along with implementation details.

\subsection{Revisiting iGPT}
\label{subsec:igpt}
\paragraph{GPT.} In NLP, the generative pretraining involves modeling the probability of the next word in a corpus $\mathcal{U} = \{u_1, ..., u_n\}$ autoregressively. This can be written as:
\begin{equation}
    p(u) = \prod\limits_{i=1}^{n} p(u_i|u_1,...,u_{i-1}, \Theta)
\end{equation}
Here, GPT computes the likelihood of each word $u_i$ based on the context of all preceding words from $u_1$ to $u_{i-1}$, aiming to minimize the negative log-likelihood of the target words:

\begin{equation}
    \mathcal{L} =  - log~ p(u)
\end{equation}

\paragraph{Image GPT.} 
In the context of images, where the input is an image $X \in \mathcal{R}^{H\times W \times C}$, the challenge lies in converting this 2D structure into a sequential format akin to a language sequence. iGPT~\cite{igpt} addresses this by na\"ively vectorizing the image $X$ into a series of individual pixels $\{x_1,...,x_n\}$, treating each pixel as analogous to a word. It then models the probability of each subsequent pixel based on the preceding ones in the sequence:

\begin{equation}
\label{eq:igpt}
    p(x) = \prod\limits_{i=1}^{n} p(x_i|x_1,...,x_{i-1}, \Theta)
\end{equation}
In this formulation, iGPT aims to predict each pixel $x_i$ utilizing the information from preceding pixels $\{x_1, ..., x_{i-1}\}$, minimizing the negative log-likelihood:

\begin{equation}
    \mathcal{L} =  - log~ p(x)
\end{equation}
Nevertheless, the extensive computational demands of iGPT, primarily due to the quadratic complexity of attention mechanisms relative to sequence length, limit its applicability for various vision tasks. For iGPT, this sequence length corresponds to the total number of pixels $Seq=H\times W$. As such, iGPT is primarily suited for low-resolution images (\eg, $Seq=32\times 32$). 

To mitigate this computational challenge, especially for high-resolution image training, approaches like SAIM~\cite{qi2023exploring} and RandSac~\cite{randsac} have been developed. A critical advancement in these methodologies is the incorporation of Vision Transformer (ViT) architecture \cite{vit}, which significantly transforms the tokenization approach --- instead of treating each pixel as an individual token, ViT redefines tokens as image patches (\eg, clusters of pixels). This strategy effectively reduces the sequence length for each image, thereby enabling the practical application of iGPT to higher-resolution images.

\subsection{D-iGPT}
\label{subsec:digpt}
Our development of D-iGPT is built upon iGPT with the ViT architecture. Unlike iGPT completely drops the knowledge of the 2D input structure, D-iGPT is designed to carefully encode this information.  Specifically, at the input level, images are divided into multiple equally-sized, non-overlapping regions, forming clusters $S = \{s_1, ..., s_n\}$. Note that each cluster contains multiple spatially neighbored image patches, and serves as a fundamental unit in the sequence for autoregressive modeling. This encoding is crucial for facilitating a more intricate interplay between different regions (rather than ``local'' patches) of an image, thereby enhancing the effectiveness of autoregressive modeling.

Consequently, the autoregressive probability, previously defined for individual pixels in iGPT (as in Equation \ref{eq:igpt}), is now reformulated for these clusters as:
\begin{equation}
\label{eq:d-igpt}
    p(s) = \prod\limits_{i=1}^{n} p(s_i|s_1,...,s_{i-1}, \Theta)
\end{equation}
By default, we configure the number of clusters to $4$, corresponding to a dimension of $112\times112$ for an input image of $224\times224$ (\eg, each cluster contains $7\times7$ images patches of the size $16\times16$), as illustrated in Figure \ref{fig:method}.

Building upon this new setup, we next introduce two simple yet essential modifications to enhance iGPT. 

\paragraph{Modification I: semantic tokens.} 
In contrast to the inherently semantically-rich nature of text, raw pixels in images generally lack such depth of meaning. Addressing this semantic discrepancy is crucial for enhancing learning efficacy in models like iGPT. To bridge this gap, our approach, inspired by BEiT \cite{BEiT}, involves transitioning the autoregressive target of D-iGPT from raw pixels to semantic tokens, which can be written as:
\begin{equation}
    \mathcal{L}_G =  -\sum\limits_{i=1}^{n} cosine(G(f(x_{s_1:s_{i-1}}); \theta_G), f_{\phi}(x)_{s_i}),
\end{equation}
where $f(\cdot)$ is the encoder, $f_{\phi}(x)_{s_{i}}$ is the semantically enriched tokens corresponding to the cluster $s_i$, $G(\cdot; ~\ \theta_G)$ is the generative decoder for autoregressive prediction, and $cosine$ is the cosine similarity loss.

Furthermore, to break the dependency on a fixed sequence order and enhance learning flexibility, we adopt strategies from \cite{hua2022self,xlnet} by randomly permuting the sequence of clusters $\{s_1,...s_n\}$ and selecting a permutation $\pi$. 

\paragraph{Mofication II: supervision on visible clusters.}

To further enhance the training of our model, we introduce additional supervision targeting visible clusters. This is formulated as:
\begin{equation}
\label{eq:d-decoder}
    \mathcal{L}_D =  -\sum\limits_{i=1}^{n} cosine(D(f(x_{s_1:s_{i-1}}); \theta_D), f_{\phi}(x)_{s_1:s_{i-1}})
\end{equation}
where $D(\cdot; ~\ \theta_D)$ is the discriminative decoder, tasked with predicting the semantic tokens of visible pixels.

This approach, as encapsulated in Equation \eqref{eq:d-decoder}, can be conceptualized as a form of knowledge distillation \cite{hinton2015distilling} --- its objective is to enable the encoder of D-iGPT (the student model) to distill knowledge from the model $f_{\phi}(x)$ (the teacher model), which provides semantic tokens, based on the visible sequence of clusters $\{s_1, ..., s_{i-1}\}$. However, our methodology differs from traditional knowledge distillation frameworks~\cite{featurekd,kd}, which typically align logits or feature maps between teacher and student models directly. Instead, we apply the knowledge distillation supervision on the separately designed discriminative decoder $D(\cdot; ~\ \theta_D)$. This design helps to disentangle different supervisions in training (\ie, autoregressive on $G(\cdot; ~\ \theta_G)$, distillation on $D(\cdot; ~\ \theta_D)$), and is crucial for ensuring the acquisition of high-quality representations, as demonstrated in the subsequent experimental section.

\paragraph{Summary.} The integration of these two modifications significantly enhances the capabilities of iGPT for visual representation learning. While there are various options for $f_{\phi}(x)$ to generate semantic tokens, our empirical findings, as detailed next, indicate a marked preference for discriminatively trained models like CLIP~\cite{clip}. 

Moreover, from an implementation perspective, we adopt the attention mask strategy as employed in \cite{gpt,igpt,randsac,gpt4}. This approach facilitates efficient computation of input sequences of varying lengths (\eg, a set of input sequences such as $\Big\{\{s_1\}$, $\{s_1, s_2\}$, ..., $\{s_1, s_2, ..., s_{n-1}\}\Big\}$) within a single iteration.  We direct interested readers to the supplementary materials for more details.

\paragraph{Architecture design.} The D-iGPT architecture is composed of two primary components: the encoder and the lightweight decoders. For the encoder, it leverages the standard ViT architecture. For the lightweight decoders, each incorporates two Transformer decoder blocks by default. 
Note that while the discriminative decoder $D$ and the generative decoder $G$ share the same architecture design, they are characterized by different sets of parameters. As shown in Figure \ref{fig:method}, they take the randomly initialized [Dis] tokens $D$ or [Gen] tokens $G$ with position information as the query, and the output features from the encoder as the key and the value.
Notably, in downstream tasks, we utilize only the encoder, discarding the decoder component.

\begin{table*}[t!]
\centering
\begin{tabular}{lcccc}
\toprule
Method & Pretraining Epochs& Tokenizer/Teacher &   Classification  &  Segmentation \\
\midrule
\multicolumn{5}{l}{\textit{Base-size models (ViT-B)}} \\
DeiT~\cite{deit}         & 300  & Label & 81.2 & 47.2\\
BEiT~\cite{BEiT}&800 & DALLE&83.2&-\\
MAE~\cite{mae}         & 1600  & Pixel&  83.6  & 48.1 \\
SdAE~\cite{SdAE} & 300  & EMA&  84.1&  48.6 \\
PeCo~\cite{peco} & 300  & VQGAN & 84.1  & 46.7 \\
TinyMIM~\cite{tinymim} & 300  &   MAE&  85.0 &  52.2\\
FD~\cite{featurekd}& 300&CLIP&84.8& -\\
BEiTv2~\cite{beitv2}&300&CLIP+VQGAN&85.0& 52.7 \\

Randsac~\cite{randsac}         & 1600  & Pixel&  83.7   & -\\
SAIM~\cite{qi2023exploring}&800&Pixel&83.9&-\\
PeCo~\cite{peco} & 800  & VQGAN&  84.5 &  48.5 \\
data2vec~\cite{data2vec} & 800  & EMA&  84.2 & - \\
SIM~\cite{sim} & 1600 & EMA  & 83.8 & - \\
iBOT~\cite{ibot} & 1600 & EMA  & 84.0 & - \\
MaskFeat~\cite{maskfeat} & 1600  & HOG& 84.0 & - \\
BEiTv2~\cite{beitv2}&1600&CLIP+VQGAN&85.5&53.1\\
DeepMIM~\cite{deepmim}&1600&CLIP&85.6&53.1\\
MILAN~\cite{MILAN}&1600&CLIP&85.6&-\\
EVA ~\cite{eva}&800&CLIP&85.5&53.3 \\
	\rowcolor{mygray}
D-iGPT (Ours)& 300&CLIP&\textbf{86.2}  &\textbf{53.8}\\
\midrule
\multicolumn{5}{l}{\textit{Large-size models (ViT-L)}} \\
BEiTv2~\cite{beitv2}&300&CLIP+VQGAN&86.6&55.0 \\
BEiT~\cite{BEiT}&800& DALLE& 85.2\\
MAE~\cite{mae}         & 1600  & Pixel& 85.9  & 53.6\\
PeCo~\cite{peco} & 800  & VQGAN&  86.5 & -\\
iBOT~\cite{ibot} & 1600 & EMA  & 84.8 & - \\
MaskFeat~\cite{maskfeat} & 1600  & HOG& 85.7 &-\\
BEiTv2~\cite{beitv2}&1600&CLIP+VQGAN&87.3&56.7 \\
MILAN~\cite{MILAN}&1600&CLIP&~86.8$\dagger$&-\\
	\rowcolor{mygray}
D-iGPT (Ours)& 300&CLIP&\textbf{87.8} & \textbf{57.3}\\
\bottomrule
\end{tabular}
\vspace{-.6em}
\caption{Fine-tuning results which methods were pretrained on \textbf{ImageNet-1K} and fine-tuned on ImageNet-1K/ADE20K on classification and semantic segmentation. $\dagger$: reproduced result using official code. }
\label{1k}
\end{table*}

\section{Experiment}

\paragraph{Implementation details.}

In our experiments, we use CLIP to provide semantic tokens. We pretrain, by default, all models on ImageNet-1K dataset for 300 epochs. We set the batch size to 4096 and the peak learning rate to $lr=1.5e^{-4}\times batchsize/256$. We adopt a cosine learning rate decay schedule with a warm-up period of 40 epochs, and utilize the AdamW~\cite{adamw} optimizer with a weight decay of 0.05. We use random resized cropping and random horizontal flipping, with the input size set to 224 $\times$ 224. 

When further scaling the pretraining to ImageNet-21K dataset, all models undergo 150 epochs of pretraining with a warm-up stage of 5 epochs, a learning rate $lr=1.5e^{-3}$, and a batch size of 4096.

\begin{table*}[t!]
\centering
\resizebox{\linewidth}{!}{
\begin{tabular}{lcccccccc}
\toprule
Method & IN-1K $\uparrow$ & IN-V2 $\uparrow$&IN-Real $\uparrow$& IN-A.$\uparrow$ &IN-Ren.$\uparrow$&IN-C. $\downarrow$ & IN-S.$\uparrow$ &IN-H. $\uparrow$\\
\midrule
\multicolumn{7}{l}{\textit{Base-size models (ViT-B)}} \\
DeiT~\cite{deit}&  81.2& 70.6& 86.7& 27.9    & 45.4 & 36.8&32.3& 23.8\\
TinyMIM~\cite{tinymim}&  85.0 & 75.3&88.7 &43.0 &54.6& 32.7& 41.0&29.2 \\
MAE~\cite{mae}   &  83.6  & 72.9&  88.1   & 33.6  &50.0& 37.8 &36.4 & 25.5  \\
BEiT~\cite{BEiT}&83.2&71.8&87.9&32.8&49.6&38.7&35.1 & 25.1\\
iBOT~\cite{ibot}&84.0&73.0&88.2&33.0&51.2&36.9&38.7 &26.3\\
BEiTv2~\cite{beitv2}&85.5&76.2&89.2&54.0&61.7&30.9&45.9& 30.2\\
	\rowcolor{mygray}
D-iGPT (Ours) &\textbf{86.2} &\textbf{76.4} &\textbf{89.6} &\textbf{56.3}&\textbf{64.3}&\textbf{29.9}& \textbf{48.5}& \textbf{31.1} \\
\midrule
\multicolumn{7}{l}{\textit{Large-size models (ViT-L)}} \\
MAE~\cite{mae}   &  85.9  & 76.5&89.4 &56.3&61.0&31.1&45.6  &32.4  \\
BEiT~\cite{BEiT}&85.2&75.1&88.8&55.4&59.8&32.0&43.8 &31.2\\
iBOT~\cite{ibot}&84.8&74.4&87.9&53.9&57.1&34.1&42.6&30.8\\
BEiTv2~\cite{beitv2}&87.3&78.3&90.0&68.6&70.3&25.4&53.7&36.5\\
	\rowcolor{mygray}
D-iGPT (Ours) & \textbf{87.8}&\textbf{79.6} & \textbf{90.4}&\textbf{73.0}&\textbf{80.5}&\textbf{24.7}& \textbf{60.3}&  \textbf{37.6}\\
\bottomrule
\end{tabular}
}
\vspace{-.8em}
\caption{Robustness and Generalization evaluation on out-of-domain datasets.}
\label{tab:robust}
\end{table*}
\subsection{ImageNet-1K Pretraining}

For a fair comparison with previous work~\cite{BEiT,beitv2,mae,maskfeat,data2vec,deepmim,peco,tinymim}, we first study pretraining on ImageNet-1K~\cite{imagenet} dataset
with ViT-B and ViT-L.

\subsubsection{ImageNet Classification}
Following \cite{mae}, we finetune pretrained models using the ImageNet-1K training set, and test on the ImageNet-1K validation set with the input size of $224\times 224$.

Note that different from previous approaches such as \cite{zhai2022scaling,yu2022coca}, which employs multi-head attention pooling, and BEiT-3 \cite{BEiT3}, which exploits an additional pretrained giant language tower as the image classification task layer, we hereby opt for a simple linear layer for classification. We finetune the pretrained model for 100 epochs.

\paragraph{Results.} 
As shown in Table \ref{1k}, 
our ViT-B impressively achieves 86.2\% top-1 accuracy. This is the first instance of a ViT-B model surpassing the 86\% accuracy threshold on ImageNet-1K, using an input size of $224\times 224$.

In terms of comparative performance, D-iGPT demonstrates a significant improvement over various existing methods. It exceeds the baseline supervised model, DeiT, by a substantial margin of \textbf{+5.0\%}, the prevalent mask image modeling method, MAE, by \textbf{+2.6\%}, and the prior art MILAN/DeepMIM by \textbf{+0.6\%}. Furthermore, with the same teacher model, D-iGPT surpasses EVA by \textbf{+0.7\%}, while requiring only 37.5\% of the training epochs.

When enlarging the model size to ViT-L, our D-iGPT sets a new benchmark with an accuracy of 87.8\%. Notably, this result surpasses the well-known mask image modeling MAE by \textbf{+1.9\%}  and prior art BEiT-v2 by \textbf{+0.5\%}.

\subsubsection{Semantic Segmentation}
For semantic segmentation, we evaluate D-iGPT using the ADE20K dataset~\cite{ade20k}, which comprises 150 categories with 20,000 training images and 2,000 validation images. Following MAE \cite{mae}, we adopt our D-iGPT pretrained ViT model as the backbone and UperNet~\cite{upernet} as the framework. The input image resolution is $512 \times 512$ for training and evaluation; we report mIoU as the evaluation metric.

The last column in Table \ref{1k} reports the performance of D-iGPT on ADE20K. We note that D-iGPT achieves a mIOU of 53.8 with ViT-B and a mIOU of 57.3 with ViT-L, which sets new benchmarks for their respective model sizes. 
These impressive results highlight the strong generalization capabilities of D-iGPT on downstream tasks. 

Additionally, we assess model robustness on out-of-domain samples. We note that D-iGPT consistently outperforms both supervised models like DeiT and self-supervised models like MAE across all out-of-domain datasets. We refer interested readers to our appendix for more details.

\subsubsection{Robustness}

We assess model robustness on various out-of-domain ImageNet datasets, including natural adversarial examples (ImageNet-A~\cite{imageneta}), semantic shifts (ImageNet-R~\cite{imagenetr}), common image corruptions (ImageNet-C~\cite{imagenetc}), image sketches (ImageNet-S~\cite{imagenets}),  ImageNet-V2~\cite{inv2}, ImageNet-Real~\cite{beyer2020we}, and ImageNet-Hard~\cite{hard}.

As indicated in Table \ref{tab:robust}, D-iGPT consistently outperforms both supervised models like DeiT and self-supervised models like MAE across all datasets, showcasing notable improvements in robustness and generalization. For example, compared with the prior art BEiT-v2, D-iGPT exhibits superior robustness with improvements ranging from 0.2\% to 2.6\% in the ViT-B model size category. These improvements are even more striking with the ViT-L model, \ie, D-iGPT makes significant strides in challenging datasets like IN-Adversarial (improvement of +4.4\%), IN-Sketch (+6.6\%), and IN-Rendition (+10.2\%).

\begin{table*}[t!]
\centering
\begin{tabular}{lccccc}
\toprule
\multirow{2}{*}{Method} &\multirow{2}{*}{Model}& \multirow{2}{*}{Model Size} & Pretraining Data& Pretraining Data&   ImageNet-1K \\
& & &   Category&  Size  &top-1 (\%) \\
\midrule

{\color[HTML]{9B9B9B} TokenLearner~\cite{ryoo2021tokenlearner}}& {\color[HTML]{9B9B9B} TokenLearner} & {\color[HTML]{9B9B9B} 460M}&{\color[HTML]{9B9B9B}~I }&{\color[HTML]{9B9B9B} 300M (Private) } &{\color[HTML]{9B9B9B} 88.9 } \\

{\color[HTML]{9B9B9B} MaxViT~\cite{tu2022maxvit}}& {\color[HTML]{9B9B9B} MaxViT} & {\color[HTML]{9B9B9B} 475M}&{\color[HTML]{9B9B9B}~I }&{\color[HTML]{9B9B9B} 300M (Private) } &{\color[HTML]{9B9B9B} 89.5 } \\

{\color[HTML]{9B9B9B} SwinV2~\cite{liu2022swin}}& {\color[HTML]{9B9B9B} SwinV2} & {\color[HTML]{9B9B9B} 3B}&{\color[HTML]{9B9B9B}~I }&{\color[HTML]{9B9B9B} 84M (Private) }  &{\color[HTML]{9B9B9B} 90.2 } \\
{\color[HTML]{9B9B9B} CoAtNet-7~\cite{dai2021coatnet}}& {\color[HTML]{9B9B9B} CoAtNet} & {\color[HTML]{9B9B9B} 2.44B}&{\color[HTML]{9B9B9B}~I }&{\color[HTML]{9B9B9B} 300M (Private) }  &{\color[HTML]{9B9B9B} 90.9 } \\
{\color[HTML]{9B9B9B} Lion~\cite{chen2023symbolic}}& {\color[HTML]{9B9B9B} ViT}& {\color[HTML]{9B9B9B} 2.44B}&{\color[HTML]{9B9B9B}~I }&{\color[HTML]{9B9B9B} 3B (Private) }&{\color[HTML]{9B9B9B} 91.1 } \\
\midrule
BEiT~\cite{BEiT}&ViT&306M &I&14M&88.6 \\
iBOT~\cite{ibot}&ViT&306M &I&14M&87.8\\
OpenClip-H~\cite{openclip}&ViT &632M& I-T&2B&88.5 \\ 
EVA~\cite{eva}&ViT &1B&I&30M  &89.6 \\ 
BEiT~\cite{BEiT}&ViT&1.9B &I-T,T,I&21M,160G,14M&89.5 \\
One-Peace~\cite{wang2023one}&Transformer&4B&I-T,A-T&2B,8k hours&89.8 \\
	\rowcolor{mygray}
D-iGPT-L (ours) &ViT&306M&I &14M  &89.5\\
	\rowcolor{mygray}
D-iGPT-H (ours) &ViT& 632M&I &14M  &90.0 \\
\bottomrule
\end{tabular}
\vspace{-1em}
\caption{Summary of D-iGPT on various vision benchmarks. I, T, and A indicate images, texts, and audios respectively. {\color[HTML]{9B9B9B} Method} indicate using private training data.}
\label{sota}
\end{table*}

\subsection{Pretraining with Larger Datasets}
Next, we explore the impact of pretraining on ImageNet-21K with \app14 million samples. Following \cite{eva,BEiT}, we initially undertake supervised fine-tuning on the ImageNet-21K training dataset for 60 epochs; subsequently, we fully finetune models on the ImageNet-1K training dataset.

\paragraph{Main Results.} The scaling results of D-iGPT, as depicted in Table \ref{sota}, are particularly noteworthy. When pretrained with ImageNet-21K, D-iGPT successfully helps ViT-L to secure a top-1 accuracy of 89.5\%. This performance not only parallels other baselines such as BEiT-3 and EVA, but also is attained with a considerably smaller model and training data size. By scaling the training to the larger ViT-H, we observe a further improvement, achieving an accuracy of 90.0\%.
This result is particularly noteworthy as it beats all existing solutions that build on public datasets; moreover, this 90.0\% accuracy is even comparable to those achieved by substantially larger models that have been trained with extensive private datasets \cite{swinv2,dai2021coatnet,chen2023symbolic}. These results demonstrate the scalability and efficacy of D-iGPT for visual representation learning.

\begin{figure}[t]
    \centering
    \includegraphics[width=0.92\linewidth]{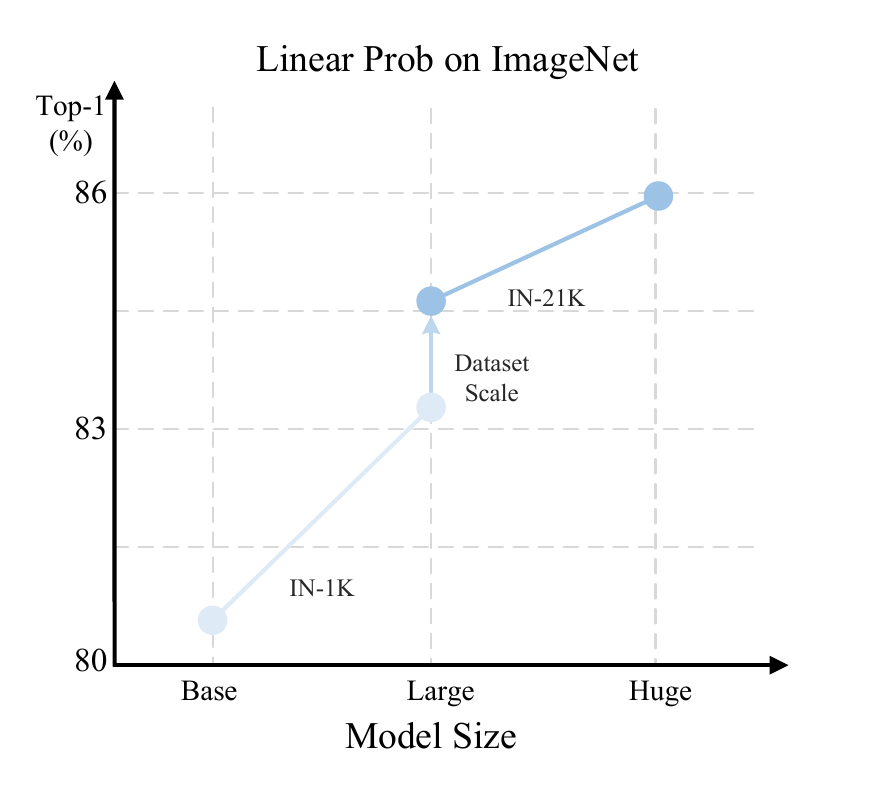}
    \vspace{-1em}
    \caption{The model and dataset scalability of D-iGPT. D-iGPT shows significant performance gain with the growth of model size and dataset size. }
    \vspace{-1em}
    \label{tab:scale}
\end{figure}

\paragraph{Linear probing.} Following \cite{beitv2,el2024scalable}, we also study model performance under the linear probing setup, \ie, we freeze the backbone and linearly evaluate the performance on ImageNet-1K. Specifically, we consider different model sizes, by scaling the vanilla ViT from Base size to Huge size, and different dataset sizes, by scaling from ImageNet-1K to ImageNet-21K.  

As shown in Figure \ref{tab:scale}, we can observe that 1) D-iGPT brings consistent improvements with bigger model sizes, and 2) larger datasets can help D-iGPT yield stronger performance. These observations demonstrate the strong scalability of D-iGPT. Additionally, it is worth mentioning that our best result is achieved by ViT-H pretrained on ImageNet-21k, with 85.9\% linear probing accuracy.

\subsection{Zero-shot Classification}

We finetune our D-iGPT on the vision-language dataset for zero-shot ImageNet classification. With such fine-tuning, our D-iGPT can be applied to a wide array of computer vision classification tasks directly with class names, without the need for task-specific fine-tuning. Additionally, the finetuned feature can be utilized in both uni-modal and multi-modal applications~\cite{llava}, akin to the capabilities demonstrated by CLIP features~\cite{clip}.

For this process, we use the D-iGPT pretrained image encoder and the OpenCLIP~\cite{openclip} pretrained text encoder as our starting point. The model is then fine-tuned on the LAION-400M dataset \cite{schuhmann2021laion,schuhmann2022laion}. The results, as summarized in Table\ref{tab:zero}, showcase significant enhancements achieved by D-iGPT. For example, compared to CLIPA \cite{clipa} and OpenClip,
D-iGPT improves the zero-shot ImageNet classification accuracy by 2.3\% and 1.8\%, respectively.
\begin{table}[t!]
    \centering
    \begin{tabular}{c|c|c|c|c}
    \toprule
        Pretraining & Model &DataSet &Samples&top-1 \\
        \midrule
         CLIPA& ViT-L/16& {\scriptsize LAION-400M}& 128M & 69.3\\
         D-iGPT& ViT-L/16&  {\scriptsize LAION-400M}& 128M & 71.6\\
         \midrule
         OpenClip& ViT-L/14& {\scriptsize LAION-400M}& 1B & 75.3\\
         D-iGPT& ViT-L/14&  {\scriptsize LAION-400M}& 1B & 77.1\\
    \midrule
    \end{tabular}
    \vspace{-1.2em}
    \caption{Zero-shot classification performance on ImageNet-1K. Samples indicate the seen samples in finetuning.}
    \vspace{-.5em}
    \label{tab:zero}
\end{table}

\subsection{Ablation Study}

\paragraph{Semantic tokens.} Our study begins with an examination of various semantic token sources. Beyond our chosen CLIP tokens and iGPT's pixel-based tokens, we also consider alternatives like DINO features~\cite{dino,maskfeat} and VQVAE tokens~\cite{beitv2}. The results, shown in Table~\ref{tab:Guidance}, reveal notable differences in performance. While autoregressive pretraining using low-level pixels or VQVAE tokens shows lesser efficacy compared to MAE, the application of tokens from discriminatively trained models significantly enhances D-iGPT's performance, surpassing MAE by a notable margin. More importantly, with CLIP as the tokenizer, D-iGPT reduces training costs by 77\% (from 666 hours to 159 hours).

\begin{table}[t]
\centering
\footnotesize
\begin{tabular}{l|c|c|c|c}
\toprule
\multirow{2}{*}{Method}  &\multirow{2}{*}{Tokenizer} & Training& ImageNet-1K& ADE20K \\
& & Cost (h)&top-1 Acc.&mIoU\\
\midrule
MAE&  Pixel    &181    & 83.2&48.0 \\
iGPT$\dagger$ &    Pixel    &80      & 82.0 \textcolor{ForestGreen}{$_{-1.2}$}&44.1 \textcolor{ForestGreen}{$_{-3.9}$}\\
\midrule
MAE&  VQVAE      &317   & 83.2 &47.2 \\
iGPT$\dagger$&         VQVAE &100& 82.3 \textcolor{ForestGreen}{$_{-0.9}$}   &47.0 \textcolor{ForestGreen}{$_{-0.2}$}   \\
\midrule
MAE&  DINO      &259   & 84.4 &50.5 \\
D-iGPT&         DINO&79  & 84.7 \textcolor{red}{$_{+0.3}$} &51.0 \textcolor{red}{$_{+0.5}$}   \\
\midrule
MAE&  CLIP    &666     & 85.4 &52.4 \\
D-iGPT&         CLIP &159&  \textbf{86.2 \textcolor{red}{$_{+0.8}$} }&\textbf{53.8 \textcolor{red}{$_{+1.4}$}}   \\
\bottomrule
\end{tabular}
\vspace{-.8em}
\caption{Ablation on different semantic tokens. $\dagger$ denotes our re-implementation with the ViT architecture. MAE is pretrained 1600 epochs while D-iGPT is pretrained 300 epochs.}
\label{tab:Guidance}
\vspace{-.5em}
\end{table}

\begin{table}[t!]
\centering
\begin{tabular}{l|c|c|c}
\toprule
\multirow{2}{*}{Student}  & Tokenizer & ImageNet-1K& ADE20K \\
&Source &top-1 Acc.& mIoU\\
\midrule

ViT-B&      CLIP-B       &    85.7   & 53.0 \\
ViT-B&      CLIP-L       &    85.9    & 53.3\\
ViT-B&      CLIP-L@336       &    84.6  & 51.8  \\
\midrule
ViT-B&     DINO-L       &   84.8 &  52.0 \\
ViT-B&      CLIP-L       &    85.9 &  53.3 \\
ViT-B&      OpenCLIP-L       &    85.9   & 53.2 \\
ViT-B&      OpenCLIP-H       &    86.2&  53.6   \\
\bottomrule
\end{tabular}
\vspace{-.5em}
\caption{Ablation on tokenizer model.}
\label{tab:teacher}
\vspace{-.5em}
\end{table}
Given the superior performance achieved with CLIP features, we next delve deeper into the effects of utilizing tokens from different CLIP variants. As detailed in Table \ref{tab:teacher}, when we use a larger tokenizer (\ie, CLIP-L), D-iGPT achieves better performance compared to using a smaller tokenizer (\ie, CLIP-B). However, interestingly, if we employ CLIP-L@336 as the tokenizer while maintaining the input size of $224\times 224$, the performance of D-iGPT drops significantly. We conjecture this is mainly due to a resolution mismatch during the training phase and the inference phase of CLIP-L@336.

Further experiments explore various large-size tokenizers, including DINO, CLIP, and OpenCLIP. One the one hand, we note that using OpenCLIP-L as the tokenizer, which is the same as CLIP-L in architecture but varies in training data, results in comparable performance to employing CLIP-L. Scaling to the even larger tokenizer, OpenCLIP-H, can further enhance D-iGPT's performance. On the other hand, we interestingly note that tokenizers like DINO do not yield comparatively favorable results. This may suggest that larger pretraining datasets and the inclusion of textual information are likely beneficial in generating high-quality semantic tokens for guiding D-iGPT's learning process.

\paragraph{Number of Clusters} We configure the number of clusters from 1 to 196, corresponding to cluster shape ranging from 224$\times$224 to 16$\times$16, as shown in Table \ref{tab:num}. The performance initially increases from 85.3\% to a peak of 86.2\% as the number of clusters increases from 1 to 4. However, further increasing the number of clusters causes the performance to gradually decline, reaching 85.4\% at 196 clusters. 
\begin{table}[t]
    \centering
    \begin{tabular}{c|c|c}
    \toprule
       Num of Clusters  & Cluster shape & Top-1(\%)\\
       \midrule
       1  & 224$\times$224& 85.3\\
       2  & 224$\times$112& 85.6\\
       \textbf{4}  & \textbf{112$\times$112}& \textbf{86.2}\\
       14  & 112$\times$32& 85.8\\
       49  &  32$\times$32&85.7\\
       196  & 16$\times$16& 85.4\\
         \bottomrule
    \end{tabular}
    \vspace{-.5em}
    \caption{Ablation study on the number of clusters.}
    \label{tab:num}
\end{table}

\paragraph{Pretraining paradigm}
In our evaluation of various pretraining paradigms, we consider Mask Image Modeling (MIM), Knowledge Distillation (KD), and our D-iGPT. To facilitate a fair comparison, especially for the MIM-based MAE model, we modify it to utilize CLIP features as the supervision target, moving away from the conventional pixel-based approach. The results are presented in Table \ref{tab:pretask}.

The baseline pretraining methods, such as MAE, EVA, and KD, exhibit comparable performance levels in both ImageNet classification and ADE20K semantic segmentation. In contrast, our D-iGPT achieves markedly better results. For instance, while the highest performance among baseline models is 85.0\% accuracy on ImageNet and 52.6 mIOU on ADE20K, D-iGPT significantly elevates these benchmarks to 86.2\% accuracy on ImageNet and 53.8 mIOU on ADE20K. These findings underscore the potential of autoregressive pretraining, implemented in D-iGPT, as a more scalable and effective visual representation learner.

\begin{table}[t]
\centering
\begin{tabular}{l|c|c}
\toprule
\multirow{2}{*}{Method}  & ImageNet-1K& ADE20K \\ &top-1 Acc.&mIoU\\
\midrule
MAE$\dagger$~\cite{mae}&   84.6&52.1 \\
EVA~\cite{eva}&    85.0 &52.6\\
KD~\cite{featurekd}&      85.0&52.5 \\
\rowcolor{mygray} D-iGPT&  86.2 &53.8   \\
\bottomrule
\end{tabular}
\vspace{-.8em}
\caption{Ablation on the pretraining paradigm. MAE$\dagger$ is our re-implementation with CLIP features as supervision targets.}
\vspace{-.5em}
\label{tab:pretask}
\end{table}

\begin{table}[t!]
\centering
\begin{tabular}{c|c|c|c}
\toprule
\multirow{2}{*}{Dec. Depth}  & \multirow{2}{*}{Dec. Dim} & ImageNet-1K& ADE20K \\
& &top-1 Acc.& mIoU\\
\midrule
1&      1024       &    85.6&  52.8   \\
2&      1024    &    86.2&  53.6   \\
4&      1024      &    86.0&  53.2   \\
\midrule
2&      512       &    85.8&  53.0   \\
2&      768    &    85.9&  53.3   \\
2&      1024      &    86.2&  53.6   \\
\bottomrule
\end{tabular}
\vspace{-.8em}
\caption{Ablation on the decoder design.}
\vspace{-1em}
\label{tab:decoder}
\end{table}

\paragraph{Decoder Design}

Our investigation begins with an examination of \emph{Decoder Depth}. Our decoder design is lightweight, with the number of layers at most 4 (by default is 2).  Intriguingly, this simpler decoder architecture not only significantly reduces GPU computational load but also enhances overall performance. As shown in Table \ref{tab:decoder}, a 2-layer decoder outperforms a 4-layer decoder, even when maintaining the same decoder dimension of 1024. In contrast, MAE, by default, uses an 8-layer decoder for attaining the best performance.

Building on the success of the 2-layer decoder, we next turn our attention to the \emph{Decoder Dimension (Dim)}. Through our experiments, we note that a reduction in decoder dimension results in a slight decrease in model performance. For example, by halving the decoder dimension from 1024 to 512, we observe an accuracy drop of 0.4\% on ImageNet and a mIOU drop of 0.6 on ADE20K. This finding highlights the nuanced impact of decoder dimensionality on D-iGPT's effectiveness.

\paragraph{Discriminative Decoder.}
We ablate the discriminative decoder that predicts the semantic
tokens of the visible pixels. Firstly, we check the setting where we remove both the discriminative decoder and the generative decoder, and implement feature distillation supervision \cite{featurekd} directly on the output feature map of the encoder. The corresponding result is reported in the second row of Table \ref{tab:dis}, showing 84.9\% accuracy. This is 1.3\% lower in accuracy than the default setup of D-iGPT (86.2\%). Besides, if we keep the generative decoder and knowledge distillation, the performance is 0.4\% and 1.2\% lower than the generative decoder only and the default setup of D-iGPT, respectively.
Next, if we keep only the discriminative decoder and remove the generative decoder, the accuracy will drop by 1.5\% (from 86.2\% to 84.7\%). This outcome underscores the critical role of the discriminative decoder in maintaining the efficacy of the pretraining process in D-iGPT.

\begin{table}[t]
\centering
\resizebox{\linewidth}{!}{
\begin{tabular}{c|c|c|c}
\toprule
\multirow{2}{*}{Method} & \multirow{2}{*}{Gen Decoder} & \multirow{2}{*}{Dis Decoder} & ImageNet-1K\\
& && (top-1 Acc.) \\
\midrule
FD~&     &        &   84.9    \\
D-iGPT&        &       $\checkmark$     & 84.7   \\
D-iGPT&       $\checkmark$ &          &   85.5 \\
D-iGPT$\dag$&       $\checkmark$ &          &   85.1 \\
D-iGPT&       $\checkmark$ &       $\checkmark$     & 86.2   \\
\bottomrule
\end{tabular}
}
\vspace{-.8em}
\caption{Ablation on the discriminative decoder. $\dag$ indicates D-iGPT takes extra distillation}
\vspace{-1em}
\label{tab:dis}
\end{table}

\section{Conclusion}
In this work, we introduce D-iGPT, an enhancement of iGPT that transitions the focus of autoregressive prediction from raw pixels to semantic tokens and supplements the supervision of visible pixels. This simple yet essential modification has led to a significant achievement: D-iGPT attains an impressive 90.0\% top-1 accuracy on the ImageNet-1K dataset, a feat accomplished using solely publicly available datasets. We hope our D-iGPT can inspire more research on rethinking autoregressive pretraining for visual representation learning and bring fresh perspectives on building vision foundation models on publicly available data sources.

\section*{Impact Statement} D-iGPT showcases its extremely strong performance on ImageNet, which can motivate the community to rethink and further explore the potential of auto-regressive pretraining for visual representation learning. Furthermore, D-iGPT potentially helps to lay the foundation for an autoregressive framework capable of universal learning with different modalities, including vision, language, audio, and more.

\subsection*{Acknowledge}
This work is supported by ONR  with N00014-23-1-2641, TPU Research Cloud (TRC) program and Google Cloud Research Credits program.

\bibliography{example_paper}

\begin{thebibliography}{66}
\providecommand{\natexlab}[1]{#1}
\providecommand{\url}[1]{\texttt{#1}}
\expandafter\ifx\csname urlstyle\endcsname\relax
  \providecommand{\doi}[1]{doi: #1}\else
  \providecommand{\doi}{doi: \begingroup \urlstyle{rm}\Url}\fi

\bibitem[Baevski et~al.(2022)Baevski, Hsu, Xu, Babu, Gu, and Auli]{data2vec}
Baevski, A., Hsu, W.-N., Xu, Q., Babu, A., Gu, J., and Auli, M.
\newblock Data2vec: A general framework for self-supervised learning in speech, vision and language.
\newblock \emph{arXiv preprint arXiv:2202.03555}, 2022.

\bibitem[Bao et~al.(2022)Bao, Dong, Piao, and Wei]{BEiT}
Bao, H., Dong, L., Piao, S., and Wei, F.
\newblock {BEiT}: {BERT} pre-training of image transformers.
\newblock In \emph{International Conference on Learning Representations}, 2022.

\bibitem[Beyer et~al.(2020)Beyer, H{\'e}naff, Kolesnikov, Zhai, and Oord]{beyer2020we}
Beyer, L., H{\'e}naff, O.~J., Kolesnikov, A., Zhai, X., and Oord, A. v.~d.
\newblock Are we done with imagenet?
\newblock \emph{arXiv preprint arXiv:2006.07159}, 2020.

\bibitem[Brown et~al.(2020)Brown, Mann, Ryder, Subbiah, Kaplan, Dhariwal, Neelakantan, Shyam, Sastry, Askell, Agarwal, Herbert-Voss, Krueger, Henighan, Child, Ramesh, Ziegler, Wu, Winter, Hesse, Chen, Sigler, Litwin, Gray, Chess, Clark, Berner, McCandlish, Radford, Sutskever, and Amodei]{gpt3}
Brown, T.~B., Mann, B., Ryder, N., Subbiah, M., Kaplan, J., Dhariwal, P., Neelakantan, A., Shyam, P., Sastry, G., Askell, A., Agarwal, S., Herbert-Voss, A., Krueger, G., Henighan, T.~J., Child, R., Ramesh, A., Ziegler, D.~M., Wu, J., Winter, C., Hesse, C., Chen, M., Sigler, E., Litwin, M., Gray, S., Chess, B., Clark, J., Berner, C., McCandlish, S., Radford, A., Sutskever, I., and Amodei, D.
\newblock Language models are few-shot learners.
\newblock \emph{ArXiv}, abs/2005.14165, 2020.

\bibitem[Caron et~al.(2021)Caron, Touvron, Misra, J\'egou, Mairal, Bojanowski, and Joulin]{dino}
Caron, M., Touvron, H., Misra, I., J\'egou, H., Mairal, J., Bojanowski, P., and Joulin, A.
\newblock Emerging properties in self-supervised vision transformers.
\newblock \emph{arXiv preprint arXiv:2104.14294}, 2021.

\bibitem[Chen et~al.(2020{\natexlab{a}})Chen, Radford, Child, Wu, Jun, Luan, and Sutskever]{igpt}
Chen, M., Radford, A., Child, R., Wu, J., Jun, H., Luan, D., and Sutskever, I.
\newblock Generative pretraining from pixels.
\newblock In III, H.~D. and Singh, A. (eds.), \emph{Proceedings of the 37th International Conference on Machine Learning}, volume 119 of \emph{Proceedings of Machine Learning Research}, pp.\  1691--1703. PMLR, 13--18 Jul 2020{\natexlab{a}}.
\newblock URL \url{http://proceedings.mlr.press/v119/chen20s.html}.

\bibitem[Chen et~al.(2020{\natexlab{b}})Chen, Kornblith, Norouzi, and Hinton]{simclr}
Chen, T., Kornblith, S., Norouzi, M., and Hinton, G.
\newblock A simple framework for contrastive learning of visual representations.
\newblock \emph{preprint arXiv:2002.05709}, 2020{\natexlab{b}}.

\bibitem[Chen et~al.(2020{\natexlab{c}})Chen, Fan, Girshick, and He]{mocov2}
Chen, X., Fan, H., Girshick, R., and He, K.
\newblock Improved baselines with momentum contrastive learning.
\newblock \emph{preprint arXiv:2003.04297}, 2020{\natexlab{c}}.

\bibitem[Chen et~al.(2023)Chen, Liang, Huang, Real, Wang, Liu, Pham, Dong, Luong, Hsieh, et~al.]{chen2023symbolic}
Chen, X., Liang, C., Huang, D., Real, E., Wang, K., Liu, Y., Pham, H., Dong, X., Luong, T., Hsieh, C.-J., et~al.
\newblock Symbolic discovery of optimization algorithms.
\newblock \emph{arXiv preprint arXiv:2302.06675}, 2023.

\bibitem[Chen et~al.(2022)Chen, Liu, Jiang, Zhang, Dai, Xiong, and Tian]{SdAE}
Chen, Y., Liu, Y., Jiang, D., Zhang, X., Dai, W., Xiong, H., and Tian, Q.
\newblock Sdae: Self-distillated masked autoencoder.
\newblock \emph{ArXiv}, abs/2208.00449, 2022.

\bibitem[Cherti et~al.(2023)Cherti, Beaumont, Wightman, Wortsman, Ilharco, Gordon, Schuhmann, Schmidt, and Jitsev]{openclip}
Cherti, M., Beaumont, R., Wightman, R., Wortsman, M., Ilharco, G., Gordon, C., Schuhmann, C., Schmidt, L., and Jitsev, J.
\newblock Reproducible scaling laws for contrastive language-image learning.
\newblock In \emph{Proceedings of the IEEE/CVF Conference on Computer Vision and Pattern Recognition}, pp.\  2818--2829, 2023.

\bibitem[Dai et~al.(2021)Dai, Liu, Le, and Tan]{dai2021coatnet}
Dai, Z., Liu, H., Le, Q.~V., and Tan, M.
\newblock Coatnet: Marrying convolution and attention for all data sizes.
\newblock \emph{Advances in neural information processing systems}, 34:\penalty0 3965--3977, 2021.

\bibitem[Dehghani et~al.(2023)Dehghani, Djolonga, Mustafa, Padlewski, Heek, Gilmer, Steiner, Caron, Geirhos, Alabdulmohsin, et~al.]{dehghani2023scaling}
Dehghani, M., Djolonga, J., Mustafa, B., Padlewski, P., Heek, J., Gilmer, J., Steiner, A.~P., Caron, M., Geirhos, R., Alabdulmohsin, I., et~al.
\newblock Scaling vision transformers to 22 billion parameters.
\newblock In \emph{International Conference on Machine Learning}, pp.\  7480--7512. PMLR, 2023.

\bibitem[Devlin et~al.(2019)Devlin, Chang, Lee, and Toutanova]{bert}
Devlin, J., Chang, M., Lee, K., and Toutanova, K.
\newblock {BERT:} pre-training of deep bidirectional transformers for language understanding.
\newblock In \emph{Proceedings of the 2019 Conference of the North American Chapter of the Association for Computational Linguistics: Human Language Technologies}, pp.\  4171--4186. Association for Computational Linguistics, 2019.

\bibitem[Dong et~al.(2021)Dong, Bao, Zhang, Chen, Zhang, Yuan, Chen, Wen, and Yu]{peco}
Dong, X., Bao, J., Zhang, T., Chen, D., Zhang, W., Yuan, L., Chen, D., Wen, F., and Yu, N.
\newblock Peco: Perceptual codebook for bert pre-training of vision transformers.
\newblock \emph{arXiv preprint arXiv:2111.12710}, 2021.

\bibitem[Dosovitskiy et~al.(2020)Dosovitskiy, Beyer, Kolesnikov, Weissenborn, Zhai, Unterthiner, Dehghani, Minderer, Heigold, Gelly, et~al.]{vit}
Dosovitskiy, A., Beyer, L., Kolesnikov, A., Weissenborn, D., Zhai, X., Unterthiner, T., Dehghani, M., Minderer, M., Heigold, G., Gelly, S., et~al.
\newblock An image is worth 16x16 words: Transformers for image recognition at scale.
\newblock \emph{preprint arXiv:2010.11929}, 2020.

\bibitem[El-Nouby et~al.(2024)El-Nouby, Klein, Zhai, Bautista, Toshev, Shankar, Susskind, and Joulin]{el2024scalable}
El-Nouby, A., Klein, M., Zhai, S., Bautista, M.~A., Toshev, A., Shankar, V., Susskind, J.~M., and Joulin, A.
\newblock Scalable pre-training of large autoregressive image models.
\newblock \emph{arXiv preprint arXiv:2401.08541}, 2024.

\bibitem[Fang et~al.(2022)Fang, Wang, Xie, Sun, Wu, Wang, Huang, Wang, and Cao]{eva}
Fang, Y., Wang, W., Xie, B., Sun, Q., Wu, L., Wang, X., Huang, T., Wang, X., and Cao, Y.
\newblock Eva: Exploring the limits of masked visual representation learning at scale.
\newblock \emph{arXiv preprint arXiv:2211.07636}, 2022.

\bibitem[He et~al.(2020)He, Fan, Wu, Xie, and Girshick]{moco}
He, K., Fan, H., Wu, Y., Xie, S., and Girshick, R.
\newblock Momentum contrast for unsupervised visual representation learning.
\newblock In \emph{CVPR}, 2020.

\bibitem[He et~al.(2022)He, Chen, Xie, Li, Doll{\'a}r, and Girshick]{mae}
He, K., Chen, X., Xie, S., Li, Y., Doll{\'a}r, P., and Girshick, R.
\newblock Masked autoencoders are scalable vision learners.
\newblock In \emph{CVPR}, 2022.

\bibitem[Hendrycks \& Dietterich(2019)Hendrycks and Dietterich]{imagenetc}
Hendrycks, D. and Dietterich, T.
\newblock Benchmarking neural network robustness to common corruptions and perturbations.
\newblock \emph{ICLR}, 2019.

\bibitem[Hendrycks et~al.(2021{\natexlab{a}})Hendrycks, Basart, Mu, Kadavath, Wang, Dorundo, Desai, Zhu, Parajuli, Guo, Song, Steinhardt, and Gilmer]{imagenetr}
Hendrycks, D., Basart, S., Mu, N., Kadavath, S., Wang, F., Dorundo, E., Desai, R., Zhu, T., Parajuli, S., Guo, M., Song, D., Steinhardt, J., and Gilmer, J.
\newblock The many faces of robustness: A critical analysis of out-of-distribution generalization.
\newblock \emph{ICCV}, 2021{\natexlab{a}}.

\bibitem[Hendrycks et~al.(2021{\natexlab{b}})Hendrycks, Zhao, Basart, Steinhardt, and Song]{imageneta}
Hendrycks, D., Zhao, K., Basart, S., Steinhardt, J., and Song, D.
\newblock Natural adversarial examples.
\newblock \emph{CVPR}, 2021{\natexlab{b}}.

\bibitem[Hinton et~al.(2015{\natexlab{a}})Hinton, Vinyals, and Dean]{kd}
Hinton, G., Vinyals, O., and Dean, J.
\newblock Distilling the knowledge in a neural network.
\newblock \emph{arXiv preprint arXiv:1503.02531}, 2015{\natexlab{a}}.

\bibitem[Hinton et~al.(2015{\natexlab{b}})Hinton, Vinyals, Dean, et~al.]{hinton2015distilling}
Hinton, G., Vinyals, O., Dean, J., et~al.
\newblock Distilling the knowledge in a neural network.
\newblock \emph{arXiv preprint arXiv:1503.02531}, 2\penalty0 (7), 2015{\natexlab{b}}.

\bibitem[Hou et~al.(2022)Hou, Sun, Chen, Xie, and Kung]{MILAN}
Hou, Z., Sun, F., Chen, Y.-K., Xie, Y., and Kung, S.~Y.
\newblock Milan: Masked image pretraining on language assisted representation.
\newblock \emph{ArXiv}, abs/2208.06049, 2022.

\bibitem[Hua et~al.(2022{\natexlab{a}})Hua, Tian, Ren, Raptis, Zhao, and Sigal]{hua2022self}
Hua, T., Tian, Y., Ren, S., Raptis, M., Zhao, H., and Sigal, L.
\newblock Self-supervision through random segments with autoregressive coding (randsac).
\newblock In \emph{The Eleventh International Conference on Learning Representations}, 2022{\natexlab{a}}.

\bibitem[Hua et~al.(2022{\natexlab{b}})Hua, Tian, Ren, Raptis, Zhao, and Sigal]{randsac}
Hua, T., Tian, Y., Ren, S., Raptis, M., Zhao, H., and Sigal, L.
\newblock Self-supervision through random segments with autoregressive coding (randsac).
\newblock In \emph{The Eleventh International Conference on Learning Representations}, 2022{\natexlab{b}}.

\bibitem[Li et~al.(2023)Li, Wang, and Xie]{clipa}
Li, X., Wang, Z., and Xie, C.
\newblock An inverse scaling law for clip training.
\newblock \emph{arXiv preprint arXiv:2305.07017}, 2023.

\bibitem[Liu et~al.(2023)Liu, Li, Wu, and Lee]{llava}
Liu, H., Li, C., Wu, Q., and Lee, Y.~J.
\newblock Visual instruction tuning.
\newblock \emph{arXiv preprint arXiv:2304.08485}, 2023.

\bibitem[Liu et~al.(2022{\natexlab{a}})Liu, Hu, Lin, Yao, Xie, Wei, Ning, Cao, Zhang, Dong, Wei, and Guo]{swinv2}
Liu, Z., Hu, H., Lin, Y., Yao, Z., Xie, Z., Wei, Y., Ning, J., Cao, Y., Zhang, Z., Dong, L., Wei, F., and Guo, B.
\newblock Swin transformer v2: Scaling up capacity and resolution.
\newblock In \emph{International Conference on Computer Vision and Pattern Recognition (CVPR)}, 2022{\natexlab{a}}.

\bibitem[Liu et~al.(2022{\natexlab{b}})Liu, Hu, Lin, Yao, Xie, Wei, Ning, Cao, Zhang, Dong, et~al.]{liu2022swin}
Liu, Z., Hu, H., Lin, Y., Yao, Z., Xie, Z., Wei, Y., Ning, J., Cao, Y., Zhang, Z., Dong, L., et~al.
\newblock Swin transformer v2: Scaling up capacity and resolution.
\newblock In \emph{Proceedings of the IEEE/CVF conference on computer vision and pattern recognition}, pp.\  12009--12019, 2022{\natexlab{b}}.

\bibitem[Loshchilov \& Hutter(2019)Loshchilov and Hutter]{adamw}
Loshchilov, I. and Hutter, F.
\newblock Decoupled weight decay regularization.
\newblock In \emph{International Conference on Learning Representations}, 2019.
\newblock URL \url{https://openreview.net/forum?id=Bkg6RiCqY7}.

\bibitem[OpenAI(2023)]{gpt4}
OpenAI.
\newblock Gpt-4 technical report.
\newblock \emph{ArXiv}, abs/2303.08774, 2023.
\newblock URL \url{https://api.semanticscholar.org/CorpusID:257532815}.

\bibitem[Peng et~al.(2022)Peng, Dong, Bao, Ye, and Wei]{beitv2}
Peng, Z., Dong, L., Bao, H., Ye, Q., and Wei, F.
\newblock {BEiT} v2: Masked image modeling with vector-quantized visual tokenizers.
\newblock \emph{arXiv preprint arXiv:2208.06366}, 2022.

\bibitem[Qi et~al.(2023)Qi, Yang, Zhu, Liu, Wu, Zhao, and Li]{qi2023exploring}
Qi, Y., Yang, F., Zhu, Y., Liu, Y., Wu, L., Zhao, R., and Li, W.
\newblock Exploring stochastic autoregressive image modeling for visual representation.
\newblock In \emph{Proceedings of the AAAI Conference on Artificial Intelligence}, volume~37, pp.\  2074--2081, 2023.

\bibitem[Radford \& Narasimhan(2018)Radford and Narasimhan]{gpt}
Radford, A. and Narasimhan, K.
\newblock Improving language understanding by generative pre-training.
\newblock 2018.

\bibitem[Radford et~al.(2021)Radford, Kim, Hallacy, Ramesh, Goh, Agarwal, Sastry, Askell, Mishkin, Clark, et~al.]{clip}
Radford, A., Kim, J.~W., Hallacy, C., Ramesh, A., Goh, G., Agarwal, S., Sastry, G., Askell, A., Mishkin, P., Clark, J., et~al.
\newblock Learning transferable visual models from natural language supervision.
\newblock In \emph{ICML}, pp.\  8748--8763. PMLR, 2021.

\bibitem[Recht et~al.(2019)Recht, Roelofs, Schmidt, and Shankar]{inv2}
Recht, B., Roelofs, R., Schmidt, L., and Shankar, V.
\newblock Do imagenet classifiers generalize to imagenet?
\newblock In \emph{International conference on machine learning}, pp.\  5389--5400. PMLR, 2019.

\bibitem[Ren et~al.(2023{\natexlab{a}})Ren, Wei, Albanie, Zhang, and Hu]{deepmim}
Ren, S., Wei, F., Albanie, S., Zhang, Z., and Hu, H.
\newblock Deepmim: Deep supervision for masked image modeling.
\newblock 2023{\natexlab{a}}.

\bibitem[Ren et~al.(2023{\natexlab{b}})Ren, Wei, Zhang, and Hu]{tinymim}
Ren, S., Wei, F., Zhang, Z., and Hu, H.
\newblock Tinymim: An empirical study of distilling mim pre-trained models.
\newblock In \emph{Proceedings of the IEEE/CVF Conference on Computer Vision and Pattern Recognition (CVPR)}, pp.\  3687--3697, June 2023{\natexlab{b}}.

\bibitem[Russakovsky et~al.(2015)Russakovsky, Deng, Su, Krause, Satheesh, Ma, Huang, Karpathy, Khosla, Bernstein, Berg, and Fei-Fei]{imagenet}
Russakovsky, O., Deng, J., Su, H., Krause, J., Satheesh, S., Ma, S., Huang, Z., Karpathy, A., Khosla, A., Bernstein, M., Berg, A.~C., and Fei-Fei, L.
\newblock Imagenet large scale visual recognition challenge.
\newblock \emph{IJCV}, 2015.

\bibitem[Ryoo et~al.(2021)Ryoo, Piergiovanni, Arnab, Dehghani, and Angelova]{ryoo2021tokenlearner}
Ryoo, M.~S., Piergiovanni, A., Arnab, A., Dehghani, M., and Angelova, A.
\newblock Tokenlearner: What can 8 learned tokens do for images and videos?
\newblock \emph{arXiv preprint arXiv:2106.11297}, 2021.

\bibitem[Schuhmann et~al.(2021)Schuhmann, Vencu, Beaumont, Kaczmarczyk, Mullis, Katta, Coombes, Jitsev, and Komatsuzaki]{schuhmann2021laion}
Schuhmann, C., Vencu, R., Beaumont, R., Kaczmarczyk, R., Mullis, C., Katta, A., Coombes, T., Jitsev, J., and Komatsuzaki, A.
\newblock Laion-400m: Open dataset of clip-filtered 400 million image-text pairs.
\newblock \emph{arXiv preprint arXiv:2111.02114}, 2021.

\bibitem[Schuhmann et~al.(2022)Schuhmann, Beaumont, Vencu, Gordon, Wightman, Cherti, Coombes, Katta, Mullis, Wortsman, et~al.]{schuhmann2022laion}
Schuhmann, C., Beaumont, R., Vencu, R., Gordon, C., Wightman, R., Cherti, M., Coombes, T., Katta, A., Mullis, C., Wortsman, M., et~al.
\newblock Laion-5b: An open large-scale dataset for training next generation image-text models.
\newblock \emph{Advances in Neural Information Processing Systems}, 35:\penalty0 25278--25294, 2022.

\bibitem[Taesiri et~al.(2023)Taesiri, Nguyen, Habchi, Bezemer, and Nguyen]{hard}
Taesiri, M.~R., Nguyen, G., Habchi, S., Bezemer, C.-P., and Nguyen, A.
\newblock Imagenet-hard: The hardest images remaining from a study of the power of zoom and spatial biases in image classification.
\newblock 2023.

\bibitem[Tao et~al.(2022)Tao, Zhu, Huang, Qiao, Wang, and Dai]{sim}
Tao, C., Zhu, X., Huang, G., Qiao, Y., Wang, X., and Dai, J.
\newblock Siamese image modeling for self-supervised vision representation learning.
\newblock \emph{arXiv preprint arXiv:2206.01204}, 2022.

\bibitem[Thoppilan et~al.(2022)Thoppilan, De~Freitas, Hall, Shazeer, Kulshreshtha, Cheng, Jin, Bos, Baker, Du, et~al.]{thoppilan2022lamda}
Thoppilan, R., De~Freitas, D., Hall, J., Shazeer, N., Kulshreshtha, A., Cheng, H.-T., Jin, A., Bos, T., Baker, L., Du, Y., et~al.
\newblock Lamda: Language models for dialog applications.
\newblock \emph{arXiv preprint arXiv:2201.08239}, 2022.

\bibitem[Touvron et~al.(2020)Touvron, Cord, Douze, Massa, Sablayrolles, and J{\'e}gou]{deit}
Touvron, H., Cord, M., Douze, M., Massa, F., Sablayrolles, A., and J{\'e}gou, H.
\newblock Training data-efficient image transformers \& distillation through attention.
\newblock \emph{preprint arXiv:2012.12877}, 2020.

\bibitem[Touvron et~al.(2023)Touvron, Martin, Stone, Albert, Almahairi, Babaei, Bashlykov, Batra, Bhargava, Bhosale, et~al.]{touvron2023llama}
Touvron, H., Martin, L., Stone, K., Albert, P., Almahairi, A., Babaei, Y., Bashlykov, N., Batra, S., Bhargava, P., Bhosale, S., et~al.
\newblock Llama 2: Open foundation and fine-tuned chat models.
\newblock \emph{arXiv preprint arXiv:2307.09288}, 2023.

\bibitem[Tu et~al.(2022)Tu, Talebi, Zhang, Yang, Milanfar, Bovik, and Li]{tu2022maxvit}
Tu, Z., Talebi, H., Zhang, H., Yang, F., Milanfar, P., Bovik, A., and Li, Y.
\newblock Maxvit: Multi-axis vision transformer.
\newblock In \emph{European conference on computer vision}, pp.\  459--479. Springer, 2022.

\bibitem[Van Den~Oord et~al.(2016)Van Den~Oord, Kalchbrenner, and Kavukcuoglu]{van2016pixel}
Van Den~Oord, A., Kalchbrenner, N., and Kavukcuoglu, K.
\newblock Pixel recurrent neural networks.
\newblock In \emph{ICML}, 2016.

\bibitem[Vaswani et~al.(2017)Vaswani, Shazeer, Parmar, Uszkoreit, Jones, Gomez, Kaiser, and Polosukhin]{vaswani2017attention}
Vaswani, A., Shazeer, N., Parmar, N., Uszkoreit, J., Jones, L., Gomez, A.~N., Kaiser, {\L}., and Polosukhin, I.
\newblock Attention is all you need.
\newblock In \emph{NeurIPS}, 2017.

\bibitem[Wang et~al.(2019)Wang, Ge, Lipton, and Xing]{imagenets}
Wang, H., Ge, S., Lipton, Z., and Xing, E.~P.
\newblock Learning robust global representations by penalizing local predictive power.
\newblock In \emph{Advances in Neural Information Processing Systems}, pp.\  10506--10518, 2019.

\bibitem[Wang et~al.(2023)Wang, Wang, Lin, Bai, Zhou, Zhou, Wang, and Zhou]{wang2023one}
Wang, P., Wang, S., Lin, J., Bai, S., Zhou, X., Zhou, J., Wang, X., and Zhou, C.
\newblock One-peace: Exploring one general representation model toward unlimited modalities.
\newblock \emph{arXiv preprint arXiv:2305.11172}, 2023.

\bibitem[Wang et~al.(2022)Wang, Bao, Dong, Bjorck, Peng, Liu, Aggarwal, Mohammed, Singhal, Som, et~al.]{BEiT3}
Wang, W., Bao, H., Dong, L., Bjorck, J., Peng, Z., Liu, Q., Aggarwal, K., Mohammed, O.~K., Singhal, S., Som, S., et~al.
\newblock Image as a foreign language: {BEiT} pretraining for all vision and vision-language tasks.
\newblock \emph{arXiv preprint arXiv:2208.10442}, 2022.

\bibitem[Wei et~al.(2021)Wei, Fan, Xie, Wu, Yuille, and Feichtenhofer]{maskfeat}
Wei, C., Fan, H., Xie, S., Wu, C.-Y., Yuille, A., and Feichtenhofer, C.
\newblock Masked feature prediction for self-supervised visual pre-training.
\newblock \emph{arXiv preprint arXiv:2112.09133}, 2021.

\bibitem[Wei et~al.(2022)Wei, Hu, Xie, Zhang, Cao, Bao, Chen, and Guo]{featurekd}
Wei, Y., Hu, H., Xie, Z., Zhang, Z., Cao, Y., Bao, J., Chen, D., and Guo, B.
\newblock Contrastive learning rivals masked image modeling in fine-tuning via feature distillation.
\newblock \emph{arXiv preprint arXiv:2205.14141}, 2022.

\bibitem[Wu et~al.(2018)Wu, Xiong, Yu, and Lin]{wu2018unsupervised}
Wu, Z., Xiong, Y., Yu, S.~X., and Lin, D.
\newblock Unsupervised feature learning via non-parametric instance discrimination.
\newblock In \emph{CVPR}, 2018.

\bibitem[Xiao et~al.(2018)Xiao, Liu, Zhou, Jiang, and Sun]{upernet}
Xiao, T., Liu, Y., Zhou, B., Jiang, Y., and Sun, J.
\newblock Unified perceptual parsing for scene understanding.
\newblock In \emph{ECCV}, 2018.

\bibitem[Xie et~al.(2022)Xie, Zhang, Cao, Lin, Bao, Yao, Dai, and Hu]{simmim}
Xie, Z., Zhang, Z., Cao, Y., Lin, Y., Bao, J., Yao, Z., Dai, Q., and Hu, H.
\newblock Simmim: A simple framework for masked image modeling.
\newblock In \emph{Proceedings of the IEEE/CVF Conference on Computer Vision and Pattern Recognition}, pp.\  9653--9663, 2022.

\bibitem[Yang et~al.(2019)Yang, Dai, Yang, Carbonell, Salakhutdinov, and Le]{xlnet}
Yang, Z., Dai, Z., Yang, Y., Carbonell, J.~G., Salakhutdinov, R., and Le, Q.~V.
\newblock {XLNet}: Generalized autoregressive pretraining for language understanding.
\newblock In \emph{Advances in Neural Information Processing Systems 32: Annual Conference on Neural Information Processing Systems 2019, NeurIPS 2019, December 8-14, 2019, Vancouver, BC, Canada}, pp.\  5754--5764, 2019.

\bibitem[Yu et~al.(2022)Yu, Wang, Vasudevan, Yeung, Seyedhosseini, and Wu]{yu2022coca}
Yu, J., Wang, Z., Vasudevan, V., Yeung, L., Seyedhosseini, M., and Wu, Y.
\newblock Coca: Contrastive captioners are image-text foundation models.
\newblock \emph{arXiv preprint arXiv:2205.01917}, 2022.

\bibitem[Zhai et~al.(2022)Zhai, Kolesnikov, Houlsby, and Beyer]{zhai2022scaling}
Zhai, X., Kolesnikov, A., Houlsby, N., and Beyer, L.
\newblock Scaling vision transformers.
\newblock In \emph{Proceedings of the IEEE/CVF Conference on Computer Vision and Pattern Recognition}, pp.\  12104--12113, 2022.

\bibitem[Zhou et~al.(2019)Zhou, Zhao, Puig, Xiao, Fidler, Barriuso, and Torralba]{ade20k}
Zhou, B., Zhao, H., Puig, X., Xiao, T., Fidler, S., Barriuso, A., and Torralba, A.
\newblock Semantic understanding of scenes through the {ADE20K} dataset.
\newblock \emph{Int. J. Comput. Vis.}, 127\penalty0 (3):\penalty0 302--321, 2019.

\bibitem[Zhou et~al.(2021)Zhou, Wei, Wang, Shen, Xie, Yuille, and Kong]{ibot}
Zhou, J., Wei, C., Wang, H., Shen, W., Xie, C., Yuille, A., and Kong, T.
\newblock ibot: Image bert pre-training with online tokenizer.
\newblock \emph{arXiv preprint arXiv:2111.07832}, 2021.

\end{thebibliography}
\bibliographystyle{icml2024}

\newpage
\appendix
\onecolumn

\section{Implementation within One Trianing Iteration}
\begin{figure}[h!]
    \centering
    \includegraphics[width=0.8\linewidth]{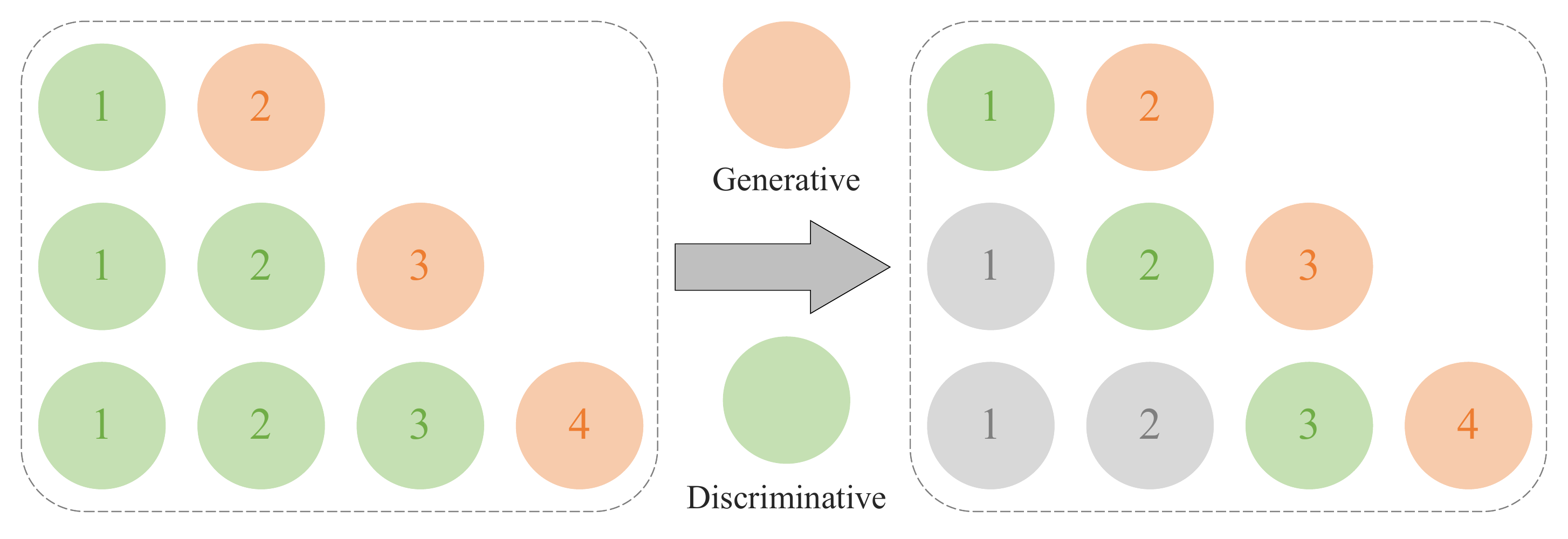}
\end{figure}

We adopt the attention mask strategy in our implementation. Specifically, for an image with 4 clusters, we can have the following $\{s_1, s_2\}$, $\{s_1, s_2, s_3\}$ and $\{s_1, s_2, s_3, s_4\}$.

By default, as shown in the left part of the figure below, the supervisions applied by the Discriminative Decoder focus on $\{s_1\}$ in $\{s_1, s_2\}$, $\{s_1, s_2\}$ in $\{s_1, s_2, s_3\}$ and $\{s_1, s_2, s_3\}$ in $\{s_1, s_2, s_3, s_4\}$. We note there are redundancies in the Discriminative Decoder's supervisions, \ie, in this single iteration, $\{s_1\}$ is being supervised for $3$ times and $\{s_2\}$ is being supervised for $2$ times.

To mitigate such redundancies, we can modify the Discriminative Decoder to supervise only on $\{s_1\}$ in $\{s_1, s_2\}$, $\{s_2\}$ in $\{s_1, s_2, s_3\}$ and $\{s_3\}$ in $\{s_1, s_2, s_3, s_4\}$ in this single iteration, as illustrated in the right part of the figure below.

To sum up, the autoregressive prediction in D-iGPT is formulated as 
\begin{equation}
    \mathcal{L}_G =  -\sum\limits_{i=1}^{n} cosine(G(f(x_{s_1:s_{i-1}}); \theta_G), f_{\phi}(x)_{s_i}),
\end{equation}
where $f(\cdot)$ is the encoder, $f_{\phi}(x)_{s_{i}}$ is the semantically enriched tokens corresponding to the cluster $s_i$, and $G(\cdot; ~\ \theta_G)$ is the generative decoder for autoregressive prediction. The supervision on visible clusters is formulated as 
\begin{equation}
\label{eq:d-decoder2}
    \mathcal{L}_D =  -\sum\limits_{i=1}^{n} cosine(D(f(x_{s_1:s_{i-1}}); \theta_D), f_{\phi}(x)_{s_{i-1}})
\end{equation}
where $D(\cdot; ~\ \theta_D)$ is the discriminative decoder, tasked with predicting the semantic tokens of visible pixels.

\end{document}